# Advice from the Oracle: Really Intelligent Information Retrieval

```
Michael J. Kurtz
Harvard-Smithsonian Center for Astrophysics
60 Garden Street, Cambridge, MA 02138 (USA)
Email: kurtz%cfazwi.decnet@cfa.harvard.edu
```

## 3.1 Introduction

What is "intelligent" information retrieval? Essentially this is asking what is intelligence, hi this article I will attempt to show some of the aspects of human intelligence, as related to information retrieval. I will do this by the device of a semi-imaginary Oracle. Every Observatory has an oracle, someone who is a distinguished scientist, has great administrative responsibilities, acts as mentor to a number of less senior people, and as trusted advisor to even the most accomplished scientists, and knows essentially everyone in the field.

In an appendix I will present a brief summary of the Statistical Factor Space method for text indexing and retrieval, and indicate how it will be used in the Astrophysics Data System Abstract Service.

## 3.2 Advice from the Oracle

### 1. The Oracle sometimes answers without being asked.

Our Oracle walks the hallways, looks into offices, and offers unsolicited advice, often. For example a conversation about the proper definition of galaxy photometry for a particular project was occurring in my office; the Oracle appeared out of nowhere and said "the amplifier on the new chip will not do any better than ten electrons readout noise", then disappeared. This was exactly the key information needed to develop a workable plan.

For machines the equivalent capability will be a long time coming, but there appear to be two central aspects to the problem. (1) How to determine which information is relevant to a problem, and (2) how to determine when relevant information is important enough to state.

To determine relevance the machine might analyze the words in our conversation, and make lists of possibly relevant enough facts. The privacy issues involved in allowing all conversations to pass through a machine which analyzes them will be skirted here.

When the machine should interrupt our conversation to interject some useful information requires rather more thought than just determining which pieces of information might be useful. Certainly having a machine constantly interrupting a conversation with relevant, but not important enough, facts would not be desirable. To be effective the machine must have a detailed knowledge of what the humans know. For example in determining an observing program for galaxy photometry the mean color of galaxies (e.g. B - R) is more important than the readout noise of the chip, but both participants in the conversation knew it.

Notice how the privacy problem is exacerbated by the machine storing what each of the participants says, and making a model for the knowledge of each participant.

## 2. The Oracle knows the environment of the problem area

Q. What photometric system should I use for this CCD study of peculiar A stars in a certain open cluster next December?

O. You should use the Stromgren system, but it is too late in the year. The objects will be too far over in the west, and because of the residual dust from Mt. Pinatubo you will not be able to get the u filter data with small enough errors to constrain your model atmosphere calculations. So wait until next year and do it in September.

Given the question even current systems, such as the ADS Abstract Service, can bring back enough relevant information to give the answer Stromgren Photometry (four of the most relevant 20 articles returned by the ADS contain Stromgren Photometry in their titles). Most of the Oracle's answer could also be gotten by a machine, but not with any existing system. One can easily imagine a rule based system to look up the air mass of a proposed observation, look up the extinction and atmospheric stability at each color, and calculate the expected error. That the atmospheric stability curve is peculiar and due to the volcano would be a note in the database containing the curve.

In order to build a system capable of duplicating the Oracle the data must exist in databases which are connected into a common system, the functions necessary to evaluate the observing conditions would need to exist, and a parser (compiler) would need to translate the question into machine instructions.

With automated literature searches and programs which simulate observations on particular instruments (e.g. on HST), the parts which remain are getting the databases connected, as ADS and ESIS are now doing, and building the rule based system. This rule based system could probably be built with no more effort than has gone into the telescope scheduling program SPIKE (Johnston, 1993). Advances in software methodologies should make this feasible by the end of the decade.

## 3. The Oracle knows what you need

Q. Should I ask for five or six nights of telescope time for this project?

O. You need to stop taking more data and write up what you've got or your thesis advisor is going to cut your stipend and boot your butt on out of here!

This is not that unusual a question. Everyone has had "Dutch Uncle" conversations like this at some time or another. One might be able to imagine information systems, perhaps first built to aid Time Allocation Committees, which could come up with this answer. What I, anyway, cannot imagine is that I would ever be willing to let a machine talk to me like that, or to accept advice of this nature from a machine. Perhaps I am just old fashioned.

## 4. The Oracle sometimes answers the question you should have asked

Q. What is the best observing strategy for a redshift survey of Abell 12345?

O. The ST group has about 300 unpublished redshifts in the can already for A12345, better choose another cluster.

This is deceptively simple. It looks like a matter of having connections to the proper databases, and knowing how to search them. Anyone who knew that another group already had 300 redshifts would not begin a cluster study.

How does the machine know this? It cannot be a simple fixed rule, imparted into the program as received wisdom by its creator. Ten years ago 30 redshifts would have been enough to stop any cluster study; now some cluster studies are undertaken despite the existence of more than 100 measurements. One could now get another 300 spectra, for a total of 600, but the bang per buck ratio would be much lower than for choosing another cluster.

The proper rule in the future cannot be predicted now with any certainty, thus we must either employ an expert human regularly updating the rules, or we must have a system complex enough to derive the rules from other data (such as the published literature). Neither is currently feasible. Making the simple conclusion, better choose another cluster. From the simple data, 300 redshifts already exist, will probably be beyond the reasoning abilities of machines for some time.

## 5. The Oracle can begin a directed dialog to get to an answer

Q. What is the best way to do photometry of spiral galaxies?

O. What do you want to use the photometry for?

Q. I want to get Tully-Fisher distances.

O. Of which galaxies?

Q. From the Zwicky catalog.

O. On the 48" use the Loral 1024 chip with the I filter, 10 minute exposures, and follow the reduction procedures of Bothun and Mould.

The ability to interact with the user is crucial for any information retrieval system, whether man or machine. The trick is asking good questions (or the equivalent). Currently machines can do this by presenting the user with a decision tree of options (e.g. menu items), by iterating on user chosen "best" answers (e.g. relevance feedback), and other means. This is a key property of intelligent systems.

## 6. The Oracle is not always right, but always respected.

Q. How should I measure velocity dispersions for elliptical galaxies.

O. Develop a system of isophotal diameters, then measure the spectrum along a long slit over the region defined by the isophotal diameter, correcting for differences in seeing.

This seems like a lot of work. One might imagine that just using a long enough slit and collapsing the spectrum to one dimension using flux weighting (i.e. doing nothing special) could work as well.

The point here is not that the Oracle can be wrong, but that the responsibility for accepting the Oracle's advice rests with the questioner. Since it is all but inconceivable that within our lifetimes a machine will issue scientific judgements which command the trust and respect which the Oracle today enjoys, it follows that very strong limits will exist for some time on what machines will be trusted to do.

## 7. The Oracle has access to more data than you know exist

Q. I plan to stop off and use the measuring engine for a week after the conference next fall.

O. Ed always takes his vacation then, and he is the only one who can fix that machine if it breaks. It would be safer to go the week before the conference.

Given a vacation schedule to replace the Oracle's educated guess of Ed's vacation plans it is not difficult to imagine a scheduling program giving this answer today. What is missing is the existence and connectivity of the databases and programs. For a ma-chine to match the Oracle a program would have to exist which could access the personnel databases to learn of Ed's work schedule, and could access the measuring engine descrip-tion, which would have to list Ed as a critical person. This program would have to be accessible to the user, and easy to use.

What the user here is doing is checking an itinerary with the Oracle. Since there are many possible places one might wish to go, and things one might wish to do once one is there, the "simple" task of knowing that Ed might be away can be seen as one of a large number of things which need to be able to be checked. Perhaps if the places one might wish to go all ran scheduling programs which could be accessed by a global scheduling program controlled by the user a capability sufficient to substitute for the Oracle could be created.

## 8. The Oracle can affect the result, not just derive it

Q. How can I make these observations when we do not have the correct set of filters?

O. If we move some money from the travel budget into the equipment budget we can afford to buy the proper filters. Please write me a one page proposal.

The Oracle is a powerful person, controlling budgets and making decisions about the allocation of resources. While one may cringe at the prospect of a machine cutting the funding for one's trip to a conference, this will certainly happen. The accounting firms are leaders in the field of information systems. Similar decisions are being made now by machine.

## 9. The Oracle has an opinion in areas of controversy

Q. What is the value of the Hubble constant?

O. 100kms"^Mpc~^, by definition.

Intelligent systems will have opinions, this is almost what we mean by intelligent systems.

Feedback loops can help the system's opinions match the user's opinions, but there will always be opinions. What these opinions are will be difficult to control, as they are the end products of complicated interactions of data with algorithms.

## 10. The Oracle often talks in riddles

Q. The grant has come through, how should I procede?

O. The great prince issues commands, founds states, vests families with fiefs. Inferior people should not be employed.

Whether the advice comes from I Ching, our Oracle, or an "intelligent" computer program it is the obligation of the questioner to interpret the information and to decide on the proper action. This remains true, even when the actual "decision" is made by the same machine which provided the information.

To the extent which key decisions are automated, either directly, or by automated decisions as to which information is relevant to bring to the attention of the human decision maker, we lose some control over our lives. For any advice/information system to be beneficial it must empower those decisions which we do make (whether by freeing time which would be wasted making less important decisions, or by providing information, or…) more than it disenfranchises us by making our decisions for us.

### 3.3 Conclusions

Machines can clearly do some things better than people. Looking through long and boring lists of measurements to select objects with the desired attributes, be they stellar spectra, galaxy images, abstracts of scientific papers, or what have you (Kurtz, 1992), is ideally suited to machines, and is what is normally referred to as "intelligent" information retrieval.

The Oracle does not compete with machines in this way. The Oracle cannot list every low surface brightness galaxy within two degrees of the Coma cluster, nor can the Oracle list every study of Lithium abundances in G stars. Machines can do these things, and will do them with ever increasing frequency.

The Oracle can combine information from a number of different sources to arrive at a conclusion, essentially what would be a vast networked system of interoperating databases exists in the head of the Oracle. Unlike the large data base case, where "intelligent" retrieval is primarily obtained by creating useful similarity measures (metrics), the Oracle sees the logical connection between groups of disparate facts, and comes to a conclusion in a way which has not been previously structured.

We have a long way to go before our machines can retrieve information in a manner as useful as the Oracle. We certainly need to get our databases connected into a common system, so that one database may be examined as a result of the examination of a different database, with these decisions made by a program. We also need to have a mechanism whereby local intelligent processes can communicate with global intelligent processes. Local processes can make use of local expertise, can access local data which could not be made globally available, and can follow local rules in disseminating information. Global processes can examine the results of a number of local processes, combine them with analyses of globally available data and other global processes, and come to conclusions which take all these things into account.

As we progress from the dawn to the first cup of coffee of the Information Age we are faced with the need to make decisions automatically in situations of great complexity. The history of chess playing programs indicates that this will not be a simple endeavor. Machines as intelligent as the Oracle are still just the stuff of science fiction, but as I can now buy a machine for $69 which can beat me at chess every time I can dream that before I die machines will be able to tell me much of what the Oracle can now. By that time the Oracle will have transformed into a Meta-Oracle, giving our progeny a new perspective on the meaning of "intelligence."

## Appendix: Statistical Factor Spaces in the Astrophysics Data System

The Astrophysics Data System Abstract Service (Kurtz et al., 1993) is a collaboration af the NASA ADS Project and the NASA Scientific and Technical Information Branch. It provides intelligent access to the NASA-STI abstract databases for astronomy and related fields. Several searching methods have been implemented in this system, each with a number of options. The user has the ability to combine the result of several different simultaneous searches in whichever way seems best. New capabilities are still being added to the system on a regular basis.

Here I will describe the Statistical Factor Space technique, a new search method which is being implemented for the first time in the ADS Abstract Service. It is based on the Factor Space technique of Ossorio (1966), but with the difference that where the original Factor Space relied on human subject matter experts to provide the basic data by filling out extensive questionnaires (essentially a psychological testing type of approach) the Statistical Factor Space obtains the basic data via an a posteriori statistical evaluation of a set of classified data (essentially building a psychometric model of the set of librarian classifiers who classified the data).

The basic data element in a Factor Space is a matrix of co-relevancy scores for term (i.e. word or phrase) versus classification (e.g. subject heading or assigned key word) pairs. Thus, for instance the term redshift is highly relevant to the subject heading cosmology, somewhat relevant to the subject heading supemovae, and not relevant to the subject heading stellar photometry. These are the measures which are assigned by humans in the original method. For the NASA STI/ADS system there are more than 10,000,000 of these items; obviously more than is feasible for humans to assign.

```
QUERY: 45.066.099 Degeneracies in Parameter Estimates for Models of
Gravitational
 Lens Systems.

 1.    45.066.204 Gravitational Lenses.
 2.    45.161.354 Light Propagation through Gravitationally Perturbed Friedmann
Universes.
 3.    45.160.045 Arcs from Gravitational Lensing.
 4.    45.066.221 Can We Measure HQ with VLBI Observations of Gravitational
Images?
 5.    45.066.012 Gravitational Lensing in a Cold Dark Matter Universe.
```

Table 3.1: The titles of the query document, and the five closest documents to it in the Factor Space. The numbers are the document numbers assigned by Astronomy and Astrophysics Abstracts.

A Statistical Factor Space is built by creating the co-occurrence matrix of term-classification pairs, and comparing each element with its expectation value were the terms distributed at random with respect to the classifications. This comparison cannot be fully parametric, as the expectation value for infrequently appearing terms is less than one, but terms can appear only in integer amounts, so even if they are distributed at random there will be cells in the co-occurrence matrix containing one, a many a variation.  The comparison function requires the co-occurrence score in a cell to be at least two before it gets weight in the decision. This systematically biases the system against infrequently occurring terms, which is exactly the opposite of the bias of the weighted word matching algorithm available in the ADS system. This suggests that the two techniques (Factor Space and weighted word matches) can be used to complement each other; the ADS system is designed to permit this easily.  The matrix of statistically derived co-relevancy scores is then manipulated using factor analysis to obtain the final relevancy space, where proximity implies nearness of meaning. Documents are then classified as the vector sum of the terms which comprise them, as are queries, and the most relevant documents to a query are simply those with the smallest Euclidean distance to the query in the relevance space. In practice the document vectors are normalized so that they occur on the surface of a unit sphere, so minimizing the Euclidean distance is equivalent to maximizing the cosine of the separation angle.

The ADS Factor Space will be built on the co-occurrence of terms with keywords assigned by the STI librarians, with the dimensionality reduced by factor analysis. A feasibility study has been successfully carried out using this technique with one half year volume of Astronomy and Astrophysics Abstracts (vol. 45), there the space was built on the co-occurrence of terms with chapter headings in A&A. Table 3.1 shows an example of how the retrieval works. The query was taken to be an abstract from the volume, number 45.066.099, titled "Degeneracies in Parameter Estimates for Models of Gravitational Lens Systems." In Table 3.1 I show the titles for the closest ^ve papers to that paper, and in Table 3.2 I show the actual terms found in the query, and in each of the closest three documents. Note in particular that the second closest document contains very few words in common, and those are not rare. It is (according to the author of the query document) a very relevant paper. Note that the term "gravitational lens" is used more than 400 times in about 100 different abstracts in the data, most of these papers not as relevant to the query paper as 45.161.354.

| 45.066.099 | 45.066.204 | 45.161.354 | 45.160.045 |
|---|---|---|---|
| cosmology | black hole | Friedmann universe | background |
| estimate | catastrophe theory | gravitational lens | clusters of galaxies |
| gravitational lens | classification | gravitational radiation | computer simulation |
| image | core | light | core |
| mass distribution | current | propagation | gravitational lens |
| mass | galaxy | | image |
| model | gravitational lens | | intergalactic medium |
| observation | law | | light |
| parameter | model | | mass distribution |
| propagation | observation | | mass |
| ray | optics | | model |
| system | position | | observation |
| time | propagation | | pair |
| transform | scaling | | symmetry |
| type | size | | |
| | star | | |
| | topology | | |

Table 3.2: The terms found in the query document in the left column, and in the three most similar documents from Table 3.1 in the next three columns.